\documentclass[letterpaper, 10 pt, conference]{ieeeconf}

\usepackage{cite}
\usepackage{amsmath,amssymb,amsfonts}
\usepackage{algorithmic}
\usepackage{graphicx}
\usepackage{textcomp}
\usepackage{xcolor}
\usepackage{mathtools}
\usepackage{authblk}
\usepackage{balance}
\usepackage[bottom]{footmisc}

\def\BibTeX{{\rm B\kern-.05em{\sc i\kern-.025em b}\kern-.08em
    T\kern-.1667em\lower.7ex\hbox{E}\kern-.125emX}}

\newtheorem{remark}{Remark}

\title{\LARGE \bf Model Predictive Control of Nonlinear Latent Force Models: \\ A Scenario-Based Approach}

\IEEEoverridecommandlockouts
\author[1]{Thomas Woodruff\textsuperscript{1}, Iman Askari\textsuperscript{2}, Guanghui Wang\textsuperscript{1} and Huazhen Fang\textsuperscript{2}
\thanks{\textsuperscript{1}T. Woodruff and G. Wang are with the Department of Electrical Engineering and Computer Science, University of
Kansas, Lawrence, KS 66045, USA (e-mail: {\tt\small tjwoodruff, ghwang@ku.edu}).}
\thanks{\textsuperscript{2}I. Askari and H. Fang are with the Department of Mechanical Engineering, University of
Kansas, Lawrence, KS 66045, USA (e-mail: {\tt\small askari, fang@ku.edu}).}
}

\begin{document}

\maketitle

\begin{abstract}
Control of  nonlinear uncertain  systems is a common challenge in the  robotics field. Nonlinear latent force models, which incorporate latent uncertainty characterized as Gaussian processes,  carry the promise of representing such systems effectively, and we focus on the control design for them in this work. To enable the design, we adopt the state-space representation of a Gaussian process to recast the nonlinear latent force model and thus build the ability to predict the future state and uncertainty concurrently. Using this feature, a stochastic model predictive control problem is  formulated. To derive a computational algorithm for the problem, we   use the scenario-based approach   to formulate a deterministic approximation of  the stochastic optimization. We evaluate the resultant scenario-based model predictive control approach through a simulation study based on motion planning of an autonomous vehicle, which shows much effectiveness. The proposed approach can find prospective use in  various other robotics applications.
\end{abstract}


\section{Introduction} 
\label{Intro}

Many robotic systems have complex nonlinear dynamics, making them challenging to be effectively controlled.  The model predictive control (MPC) approach has shown increasing success in various robot control tasks, thanks to its capability of performing predictive optimization under operating constraints~\cite{Rosolia:ARCRAS:2018}. Conventional MPC design requires an accurate model so as to predict a system's behavior over a receding horizon. This need, however, is often unsatisfiable in the real world, since robots are generally subjected to uncertainty. This uncertainty can result from unmodeled dynamics, model errors, or external disturbances. The consequences include poor control performance and even compromised safety due to constraint violation. Robust MPC techniques have thus been pursued. Among them, tube-based and min-max MPC considers set-bounded uncertainty and  seeks to keep a system's dynamic behavior bounded or optimized in the worst case, respectively~\cite{Saltik:JPC:2018,Bemporad:Springer:1999}. Other important methods include stochastic MPC, designed to handle stochastic uncertainty, and adaptive MPC, to deal with parametric uncertainty~\cite{Heirung:CCE:2018,Zhang:AUTO:2020,Bujarbaruah:arXiv:2018}. 

Gaussian processes (GPs) have emerged recently  as a useful means to model the uncertainty affecting  a dynamic system. A GP is a stochastic process in which any collection of the constituent random variables follows a multivariate Gaussian probability distribution~\cite{Rasmussen:MITPress:2005}. It provides several main advantages in describing uncertainty. First, GPs are fit to capture a broad range of uncertainties, with their probabilistic nature and ability to account for the temporal or spatio-temporal correlation. As such, they are more expansive  than  the popular  set-bounded and  random i.i.d. uncertainty models. Second, as a machine learning approach, GPs give a formalized representation of uncertainty by learning from data. This allows them to explicitly predict future uncertainty, to the benefit of control design. Note that the uncertainty prediction that GPs can offer is especially promising for MPC to enhance the control performance. 

GP-based MPC  has attracted a growing amount of research in the past years. In general, the studies consider dynamic models supplemented by GP-based uncertainty terms, which are referred to as {\em latent force models} (LFMs). MPC for a linear LFM  is developed in~\cite{Klenske:TCST:2016}, which incorporates GP-based prediction of a periodic disturbance into the receding predictive optimization. More recent studies  have shown that GP-based MPC  can be designed to deal with more complicated problems, e.g., those involving nonlinear dynamics and state-dependent disturbances, with application to  autonomous racing and path tracking by robotic arms and mobile robots~\cite{Ostafew:JFR:2016,Carron:RAL:2019, Hewing:TCST:2020}. Here, we focus only on time-correlated uncertainty. Meanwhile, GPs can also be exploited to model not only the uncertainty, but also the entire system's nonlinear dynamics, and MPC design in this context is investigated in~\cite{Kocijan:ACC:2004,Cao:JIRS:2017,Yang:SYSID:2015}.  
 
The use of  GPs, however, is data-driven and   demands considerable memory and computational power. To alleviate  this limitation, MPC with sparse GP regression has been explored in~\cite{Carron:RAL:2019, Hewing:TCST:2020,Kim:IFAC:2017}. An arguably more efficient way  is    to use   equivalent state-space representations  of   GPs. This facilitates a simpler integration of GPs into LFMs, making them more amenable to control design. Along this line, the work  in~\cite{Sarkka:TAC:2019} performs linear quadratic regulator design, and further, an MPC method  is proposed in~\cite{Grasshoff:CDC:2019}. However, these two studies consider only linear LFMs, even though practical robotic systems are   better described as nonlinear LFMs. 

Control of nonlinear LFMs is non-trivial. To deal with this open problem, we consider GP-based   uncertainty,  either additive or multiplicative, and characterize it as a state-space model. The nonlinearities and stochasticity of the model and disturbance pose obvious challenges, but a further challenge is the fact that the LFM state-space is uncontrollable~\cite{Sarkka:TAC:2019}. We   propose a stochastic MPC formulation for the augmented system representation and adopt a scenario-based approach to address the stochastic optimization by converting it into a deterministic optimization problem. Our approach is efficient and easy-to-implement by design. We  apply it to motion planning for an autonomous vehicle and offer a detailed evaluation. 

The rest of this paper is as follows: Section \ref{LFM} lays out the preliminaries about nonlinear LFMs.   Section \ref{MPC} develops and tackles  stochastic MPC   for a nonlinear LFM. In Section \ref{Results}, we apply the proposed approach to an autonomous driving simulation example. Finally, Section \ref{Conclusions} gathers concluding remarks.

\section{Overview of Nonlinear LFMs}
\label{LFM}

In this section, we begin with a brief overview of GPs, which is mainly taken from~\cite{Rasmussen:MITPress:2005}, and then introduce nonlinear LFMs involving a GP-based disturbance. 

\subsection{Preliminaries of GPs} 

GPs are a  machine learning model to approximate  complex nonlinear functions through computing the probability distributions  underlying   data generated from a function. 
They have gained increasing popularity in various data-driven tasks and have recently shown promise for modeling uncertainty inherent to dynamic systems. 

Specifically, a GP is defined as a stochastic process  such that a finite collection of the random variables are jointly Gaussian. Consider a temporal  function $w(t)$. A GP defined over $w(t)$ is denoted as follows:
\begin{subequations}\label{GP-model}
\begin{align}
w(t) &\sim \mathcal{GP} \left(m (t;\theta), \kappa(t, t^\prime;\theta) \right),\\
m(t;\theta) &= \mathbb{E} \left[ w(t) \right],\\
\kappa(t, t^\prime;\theta)  &= \mathbb{E} \left[
\left(w (t) - m(t) \right) \left(w (t^\prime) - m(t^\prime) \right) 
\right],
\end{align}
\end{subequations}
where $m$ is the mean function,   $\kappa$ is the covariance function, and $\theta$ is the hyperparameter vector.  The covariance function $\kappa$ specifies the temporal correlation  between pairs of random variables, and its selection encodes prior knowledge about $w(t)$, such as   periodicity or smoothness. Without loss of generality, we assume $m(t)=0$.  Suppose that we collect $\left \{ w(t_1), w(t_2), \ldots, w(t_N) \right \}$ and denote it as $\tilde w_N$. Based on~\eqref{GP-model}, $\tilde w_N$ will follow a joint Gaussian distribution:
\begin{align*}
\tilde w_N \sim \mathcal{N} \left(
0, K_{N N}
\right),
\end{align*}
where the $ij$-th element of $K_{N N}$ is $\kappa(t_i, t_j; \theta)$. This offers a probabilistic representation to model the data $\tilde w_N$. Provided the GP model and $\tilde w_N$, we now intend to predict the distribution of $w(t_*)$ at  time   $t_*$, which is known as a regression or prediction problem. To achieve this, let us consider the joint distribution of $\tilde w_N$ and $w(t_*)$, which is
\begin{align*}
\left[
\begin{matrix}
\tilde w_N \cr w(t_*)
\end{matrix}
\right] \sim
\mathcal{N} \left( 
\left[
\begin{matrix}
0 \cr 0
\end{matrix}
\right], 
\left[
\begin{matrix}
K_{N N} & K_{N *} \cr K_{N *}^\top & K_{* *}
\end{matrix}
\right]
\right),
\end{align*}
where the $i$-th element of $K_{N *}$ is $\kappa(t_i, t_*; \theta)$, and $K_{* *} = \kappa(t_*,t_*; \theta)$. If $K_{NN}$ is a positive definite matrix, we have
\begin{align}\label{GP-regression}
w(t_*) \, | \, \tilde w_N \sim \mathcal{N} \left(
K_{N *}^\top K_{N N}^{-1} \tilde w_N, K_{* *}-K_{N *}^\top K_{N N}^{-1} K_{N *}
 \right),
\end{align}  
according to the marginal Gaussian distribution property. Note that~\eqref{GP-regression}  gives the predictive mean and covariance of $w(t_*)$ conditioned on the observed data.  

The above shows that GPs offer a   non-parametric and probabilistic approach for data-driven modeling and prediction. Therefore, they can be   used to  describe different kinds of disturbances that may affect a dynamic system in practice. Furthermore, once learned from data, they   can predict the disturbance in the future, thus  allowing for more effective control to suppress the effect of the disturbance.  

\subsection{Nonlinear LFMs Using GPs} 

Consider a nonlinear  LFM of the form
\begin{equation}
    \label{dynamics}
	dx(t) = f(x(t), u(t), w(t), t) dt ,
\end{equation}
where $ x \in \mathbb{R}^{n_x}$ is the   state, $u  \in \mathbb{R}^{n_u}$ is the control input,   $w  \in \mathbb{R}^{n_w}$ is the latent disturbance, and $f$ is a nonlinear mapping. Here, $w(t)$ is a zero-mean GP with  $w(t) \sim \mathcal{GP}(0,\kappa(t,t'; \theta))$.  

Control design for the above nonlinear LFM entails  the  challenge of formulating and solving a control problem that can accommodate the GP-based disturbance $w(t)$. A particular difficulty lies in how to combine the data-driven GP with the state-space model in \eqref{dynamics}. Adding to this,   GP regression, as shown in~\eqref{GP-regression}, involves the use of all available data points, causing high computational complexity. To overcome this challenge, we first need to represent the GP in a   way conducive to control design. A valuable and principled approach in this regard is to exploit the state-space representation of a Gaussian process, which is outlined below.

As shown in ~\cite{Hartikainen:MLSP:2010}, a GP can be equivalently expressed as the output of a linear time-invariant stochastic differential equation driven by a white noise process, if its covariance function   is stationary, i.e., $\kappa(t,t'; \theta)=\kappa(t-t'; \theta)$. 
Hence, $w(t)$ can be written as a state-space model:  
\begin{subequations}
    	\label{GPSS}
    \begin{align}
    	dz (t) &= A z (t)dt + B  d\beta (t) \\
    	w (t) &= C z (t),
    \end{align} 
    \end{subequations}
where $z$ is an intermediate state vector defined as $z  = \left[w  \  {dw}/{dt} \ \dots \  {d^{p -1}w}/{dt^{p -1}} \right]^\top \in \mathbb{R}^{p}$, and $d\beta(t)$ is a white noise process with spectral density $Q_\beta$. The stationarity assumption is not restrictive in practice as many realistic disturbances, such as wind gusts or uneven terrain, are stationary. To explain~\eqref{GPSS} further, we demonstrate how to obtain its explicit form in the special case of   a scalar disturbance $w(t)$. First, we can use spectral factorization to factor the spectral density of $\kappa(t,t')$, $S (\omega)$,  as follows:
\begin{equation}
    \label{covSpecDense}
    S (\omega) = H(i\omega) q H(-i\omega).
\end{equation}  
Then, $H$ will represent the transfer function of~\eqref{GPSS} and $q$ represents the spectral density of the driving white noise process. The system matrices in this case are further given by
\begin{align*}
    \label{GPSSmatrix}
	A  & = 
	\begin{bmatrix}
    0 & 1 & & \\
     & \ddots & \ddots &  \\
     & & 0 & 1 \\
    -a_{ 0} & \dots & -a_{ p-2} & -a_{ p-1}
	\end{bmatrix}
, \ 
	B  = 
	\begin{bmatrix}
     0\\
    \vdots \\
    0 \\
    1
	\end{bmatrix}
,   \\
    \\
    C  &= [1, 0, \dots, 0],
\end{align*}
where the coefficients $a_i$ are the roots of  $H$. The noise spectral density $Q_\beta$ is constructed using $q$. In order for the output of the SDE to exactly represent a GP, the spectral density in \eqref{covSpecDense} must be rational and the denominator must be a polynomial in $\omega^2$~\cite{Hartikainen:MLSP:2010}. The choice of the covariance function plays a significant role in accurately representing the physical phenomena as well as determining the dimension $p$ of the latent state-space. Common choices that have satisfactory spectral densities are the Mat{\'e}rn, periodic, and polynomial among others. The popular squared exponential is irrational requiring an infinite dimensional state-space. Therefore, a truncated approximation using Taylor Series or Pad\'{e} approximants must be found instead ~\cite{Hartikainen:MLSP:2010, Sarkka:MLSP:2014}. The representation shown in \eqref{GPSS} allows a GP to be seamlessly incorporated into the original nonlinear latent force model, thus paving the way for control design. 

Based on \eqref{GPSS}, the original nonlinear LFM in \eqref{dynamics} can be converted into an augmented state-space model. Specifically, defining $\bar x (t) = \left[x^\top(t) \ z^\top(t) \right]^\top \in \mathbb{R}^{n_a}$ for $n_a = n_x +   p $, we have
\begin{equation}
	\label{AugSysTh}
	d \bar x (t) = \bar  f  \left(\bar x(t), u(t), t \right) dt + \bar B  d\beta (t), \\
\end{equation}
where  $\bar f$ and $\bar B$ are concatenations of the physical and latent state-space systems. 
For~\eqref{dynamics}, $x(t)$ may not be directly accessible in practice; instead, measurements can be made of the dynamic process. Hence, \eqref{AugSysTh} is supplemented by a measurement model given by
\begin{equation}
	\label{MeasureTh}
	y(t_k) = h(\bar x (t_k)) + v(t_k), 
\end{equation}
where $y$ is the measurement,  $ t_k$ is the discrete time index, and   $v$ is  a white Gaussian measurement noise. 
Given \eqref{AugSysTh} and \eqref{MeasureTh}, the GP hyperparameters, $\theta$, can be learned. It is important to note that the parameters of the GP are learned using the measurements of the augmented state which contains information about $\theta$. This is conducive for practical control tasks. When the disturbance can be accessed directly, as in the case of full state feedback, learning can be done by optimizing the log marginal likelihood as in standard GP regression practice~\cite{Rasmussen:MITPress:2005}. For nonlinear measurement functions, estimating the parameters becomes more difficult, though still possible, as noted in~\cite{Hartikainen:ICML:2012}. After the hyperparameters of the GP have been learned, $\bar{x}(t)$ can be estimated using nonlinear estimation techniques such as moment matching or particle filtering~\cite{Zhao:arxiv:2020, Andrieu:bookch:2001}.

\section{MPC for Nonlinear LFM} 

In this section, we formulate a stochastic MPC problem to control  the nonlinear LFM and then propose  a scenario-based approach to solve it. 
\label{MPC}
\subsection{Problem Formulation} 
To start with, we apply the Euler-Maruyama discretization method to~\eqref{AugSysTh} and obtain its discrete-time version:
\begin{align}
	\label{DiscAugSys}
	\bar x_{k+1} =  \bar x_k + \bar f( \bar x_k, u_k) \Delta t + \bar  B d\beta_k,
\end{align}
where $\Delta t$ is the sampling period, and the subscript $k$ denotes the discrete time index for notational simplicity.  In addition, $d\beta_k$ is a Gaussian white noise process with covariance   $Q_\beta \Delta t$. 

Since many practical dynamic systems are subject to control and state constraints, we need to impose such constraints when applying the   model in~\eqref{DiscAugSys}. The control input is assumed to lie in a bounded set $u_k \in \mathbb{U}$, where
\begin{align*}
	\mathbb{U} \coloneqq \{u \; \rvert \; g_u(u) \leq 0\},
\end{align*}
where $g_u$ can be a combination of linear or nonlinear functions to characterize the input constraints. 
In the case of autonomous driving, for example, the steering angle and acceleration are upper bounded as control input variables. 
The state $x_k$ is also subject to a set of constraints represented by
\begin{align*}
	\mathbb{X} \coloneqq \{x \; \rvert \; g_x(x) \leq 0\},
\end{align*}
where $g_x $ is  a collection of linear and nonlinear functions. Because of the stochastic nature of~\eqref{DiscAugSys}, it is impossible to always ensure $x_k \in \mathbb{X}$. We therefore enforce   chance-based constraints on $x_k$:
\begin{align*}
	\text{Pr} \left( x_k \in \mathbb{X} \right) \geq 1 - \epsilon,
\end{align*}
where $\epsilon$ represents the constraint violation probability and serves as a design parameter in practice. 

To control the original nonlinear LFM in~\eqref{dynamics}, we intend to leverage the predictive capability of~\eqref{DiscAugSys} and  design a stochastic  MPC controller. It is formulated as follows:
\begin{subequations}
\label{MPCbasic}
	\begin{align}
	\min_{u_0, \dotsc, u_{N-1}} \quad &   \mathbb{E} \left\{ \sum_{k=0}^{N-1} c(   x_k,u_k) + c_f(x_N ) \right\}, \label{bcost}\\ 
	\text{s.t.} \quad  &  \bar{x}_{k+1} = \bar{x}_k + \bar f(\bar{x}_k, u_k) \Delta t + \bar B  d\beta_k, \label{bdyn}\\
	 &\mathbb{E}(\bar{x}_0)= \hat {\bar x}_0 , \label{binit}\\
	 &\text{Pr} \left( x_k \in \mathbb{X} \right) \geq 1 - \epsilon, \label{bstconst} \\
	 &u_k \in \mathbb{U} \label{bctlconst}, \\
	\nonumber & k=0,1,\ldots, N-1.
	\end{align}
\end{subequations}
In the above, the stage cost $c $ relies only on $x_k$ and $u_k$  since $z_k$ is uncontrollable~\cite{Sarkka:TAC:2019}, and $c_f$ is the terminal cost. At the initial time, $\bar x_0$ is estimated  as $\hat{\bar x}_0$, which can be obtained by nonlinear state estimation techniques as previously suggested. It is seen that~\eqref{MPCbasic} presents a finite-horizon stochastic optimal control problem. At every time instant, a control input sequence, $\{u_0^* \dots u_{N-1}^* \}$, is obtained by solving this problem, and only the first element $ u_0^*$ is applied    to control the system, with the rest discarded. This procedure repeats itself  through time    in a receding-horizon fashion.  The above stochastic MPC problem is in general difficult to solve due to the uncertainty propagation and probabilistic state constraints. We leverage a scenario-based approach to address the above MPC problem. 

\subsection{The Scenario Approach} 
The scenario approach is a sampling-based strategy that handles optimization-based control of uncertain systems by creating \textit{scenarios} of an uncertainty variable sampled from its probability distribution~\cite{Calafiore:TAC:2006}.  Specifically, a scenario represents a sequence of   samples of a considered uncertainty over the prediction horizon. Based on the  scenario, one can realize  a deterministic trajectory of the states over the prediction horizon. This allows us to convert the stochastic optimization in \eqref{MPCbasic} to a deterministic problem. Including only one scenario provides only one sample trajectory for the optimization. To increase the robustness of the solution, multiple scenarios can be created such that the optimization is computed over all of them. Using more scenarios in the optimization provides a better portrayal of the uncertainty's possible effect on the system, and therefore affords the controller more information to counteract the disturbance. The scenario approach also provides a probabilistic sense of constraint satisfaction as it effectively approximates an infinite number of constraints with a finite number of constraints. The scenario approach provides a  convenient and effective method to tackle stochastic optimization and control, and has shown much efficacy for dealing with stochastic MPC~\cite{Bernardini:CDC:2009,Schildbach:AUTO:2014,Shang:JPC:2019,Kang:ACC:2015}. 
Here, we exploit it to address the problem in~\eqref{MPCbasic}. 

We generate $N_s$ scenarios of  $d\beta_k$ in~\eqref{DiscAugSys} by sampling from its assumed Gaussian distribution, as it now represents the only major source of uncertainty in the system. The $i$-th scenario is denoted as $\left\{ d\beta_k^{(i)}, k=0,\ldots,N-1 \right\}$. All the scenarios are then included into the dynamics in~\eqref{DiscAugSys}  across the entire prediction horizon, which lead  to different realizations of the system's behavior. Aggregating these realizations, we can convert the stochastic optimization in~\eqref{MPCbasic} to a deterministic problem:
\begin{subequations}
	\label{MPCScen}
	\begin{align}
	 \min_{u_0, \dotsc, u_{N-1}} \quad & \sum_{i=1}^{N_s} \sum_{k=0}^{N-1} c(x_k^{(i)},u_k) + c_f(x_N^{(i)}), \\ 
	\text{s.t.} \quad &\bar{x}_{k+1}^{(i)} = \bar{x}_k^{(i)} + \bar f(\bar{x}_k^{(i)},u_k)\Delta t +  \bar B  d\beta_k^{(i)}, \\
	 &\bar{x}_{0}^{(i)}=\hat{\bar x}_0,  \\
	 &x_k^{(i)} \in \mathbb{X}, \\
	 &u_k \in \mathbb{U}, \\
	\nonumber & k=0, \ldots, N-1, \ i=1, \ldots, N_s.
	\end{align}
\end{subequations}
The above scenario-based MPC seeks to compute the optimal control actions over all the scenarios, while respecting the state and input constraints. It can be implemented using deterministic constrained optimization solvers, e.g., those based on sequential quadratic programming or interior point methods.   In general, the method is relatively easy to implement. The scenarios are able to propagate the effect of the disturbance over the horizon and provide two key benefits. The expectation in the value function is approximated through the sampling based approach, and the constraints are easily included because of the deterministic nature of the problem.  Notice also that the computation of the optimization scales linearly with the number of scenarios used.

\begin{remark}
The formulation  in \eqref{MPCScen} considers only the uncertainty due to $d\beta_k$, assuming that $\hat {\bar x}_0$ is accurate. However, it can be readily extended to the case when the state estimation  involves much uncertainty. An approach, as suggested in~\cite{Sehr:IFAC:2017}, is to construct scenarios by sampling from the probability distribution of $\hat{ \bar{x}}_0$ in addition to that of $d\beta_k$ and then synthesize  a scenario-based optimization problem along similar lines in~\eqref{MPCScen}.  
\end{remark}

\begin{remark}
A practical question arising in  the implementation of~\eqref{MPCScen}  is the number of scenarios that should be created. For linear systems, there exist results that relate the needed number of scenarios with the constraint violation probability $\epsilon$~\cite{Campi:SIAM:2008}. Such a relation yet remains unavailable for nonlinear systems and requires further research. It should also be noted that, even though the constraints will be violated less when more scenarios are used, the optimization may become less feasible and more conservative. Hence, one needs to strike a tradeoff, and a sensible way in practice is to empirically find an appropriate number of scenarios via fine tuning for a given application. 
\end{remark}

\section{Motion Planning Case Study} 
\label{Results}
To evaluate the performance of the controller we apply it to an autonomous driving example, specifically a motion planning problem\footnote{Code can be found at https://github.com/KU-ISSL/MPC-NLFM-Scenario-ICRA21}. We start with our nominal dynamics, adopted from a kinematic bicycle model in~\cite{Rajamani:book:2011}, given by 
\begin{equation}
    \label{VehSys}
    \begin{bmatrix}
    dp_x \\
    dp_y \\
    dv \\
    d\psi
    \end{bmatrix}
    =
    \begin{bmatrix}
    v(t) \cos\Big(\psi(t) + \alpha \big(\delta(t)\big)\Big) \\
    v(t) \sin\Big(\psi(t) + \alpha \big(\delta(t)\big)\Big) \\
    a(t) + w(t) \\
    \frac{v(t)}{l/2} \sin\Big(\alpha \big(\delta(t)\big)\Big)
    \end{bmatrix}
    dt .
\end{equation}

\begin{figure*}[ht!]
\begin{center}
\includegraphics[trim={3cm 6.5cm 2.75cm 7cm},clip,width=\textwidth]{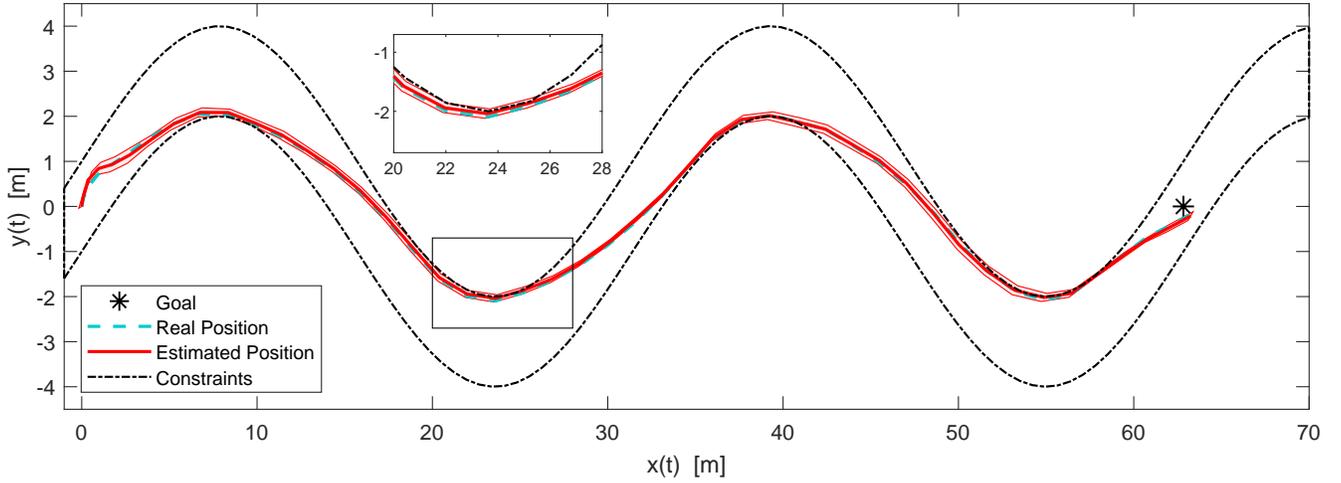}
\end{center}
\caption{Planned trajectory for autonomous driving based on the proposed approach. The true x- and y-positions are shown by the green dashed line, and their estimate and $2\sigma$ bounds is shown by the red line. The road boundary constraints are shown by the dashed black lines. Note that a small region surrounding the goal point was used as a sufficient stopping point for implementation feasibility.}
\label{fig:Position}
\end{figure*}

The state is given by its position, velocity, and heading $[p_x\ p_y\ v\ \psi]^\top$, and there are two control inputs for the acceleration and steering angle, which are denoted as $[a \ \delta]^\top$. The slip angle, which accounts for the angular difference between the velocity vector and the heading angle, is governed by $\alpha(\delta) = \arctan(0.5 \tan(\delta))$. The parameter $l$ is the length of the car and is taken to be $0.5\:\text{m}$. The dynamics are discretized according to \eqref{DiscAugSys} with a sampling time of $\Delta t = 0.2\text{s}$. A single latent disturbance is added to the velocity state and assumes a Mat{\'e}rn covariance function given by 
\begin{equation}
\begin{split}
\label{MaternCov}
\kappa_{\text{Mat{\'e}rn}}(t-t') = \sigma^2 \frac{2^{1-\nu}}{\Gamma(\nu)} \left(\frac{\sqrt{2\nu} \; (t-t')}{\ell} \right)^\nu \\
 \;  K_\nu\left(\frac{\sqrt{2\nu} \; (t-t')}{\ell} \right)
\end{split}
\end{equation}
where $\sigma$ and $\ell$ are the scaling factor and length scale, respectively, and $\nu$ is a smoothing parameter. These parameters are set as $\sigma^2 = 4$, $\ell= 4$, and $\nu = 5/2$, and provided to the controller a priori. In the above, $\Gamma(\nu)$ is the Gamma function, and $K_\nu(\cdot)$ is the modified Bessel function. The Mat{\'e}rn kernel is a useful function in that it is very general such that for $\nu = 1/2$ it is equivalent to the exponential kernel, and for $\nu = \infty$ it is equivalent to the squared exponential kernel. Its spectral density can also be factored such that the spectral factorization process produces exactly the desired GP output. We utilize a bootstrap particle filter to estimate the state, where the measurements are full-state feedback for the physical states only. The number of particles used is $N_p = 7000$. This is somewhat a large number of particles required for the estimation. This is due in part by the nonlinearities of the system as well as the fact that the importance distribution is taken to be the transitional distribution, which can lead to poor sample efficiency. More particles are also required because of the increased state dimension from the added disturbance states. More sophisticated methods could be used to decrease the required number of particles.

In the motion planning problem, the controller aims to plan an optimal trajectory to a given goal state $x_g$ under specified constraints. To do this, we construct the following cost function:
\begin{subequations}
\begin{align}
    \label{QuadCost}
    c(x_k,u_k) &= (x_k - x_g)^T Q (x_k - x_g) + u_k^T R u_k \\
   c_f(x_N) &= (x_N - x_g)^T Q_f (x_N - x_g), 
\end{align}
\end{subequations}
where the matrices $Q \succ 0$ and $Q_f \succ 0$ weight the distance from the current state to the goal, and $R \succ 0$ weights the control inputs. The weights are given by $Q=Q_f=\text{diag}(2, 2, 1, 0)$ and $R=\text{diag}(0.5, 10)$ and they serve as tuning parameters to improve overall performance. The cost function is considered over a horizon length $N=7$. During the motion planning, the vehicle must stay within the bounds of the given constraints. The track constraints are generated by a sinusoidal path in the $y$-direction and the $x$-direction is given simple bounds. Constraints for the velocity state are given by $0\:\text{m/s} \leq v \leq 8\:\text{m/s}$, and the two control inputs are constrained by $|a| \leq 5\:\text{m/}\text{s}^2$ and $|\delta| \leq 25^{\circ}$. Utilizing the cost function in \eqref{QuadCost} and the constraints provided, we construct the optimization problem. Following the scenario approach, we generate a sufficient number of scenarios, $N_s=150$, and use them in the new deterministic optimization problem. To optimize the nonlinear cost under constraints, we use the MATLAB function \textit{fmincon} with the interior points method~\cite{MATLAB:2019}. 

\begin{figure}[t]
\begin{center}
\includegraphics[trim={4.25cm 8.5cm 4.25cm 8.75cm},clip,width=\linewidth]{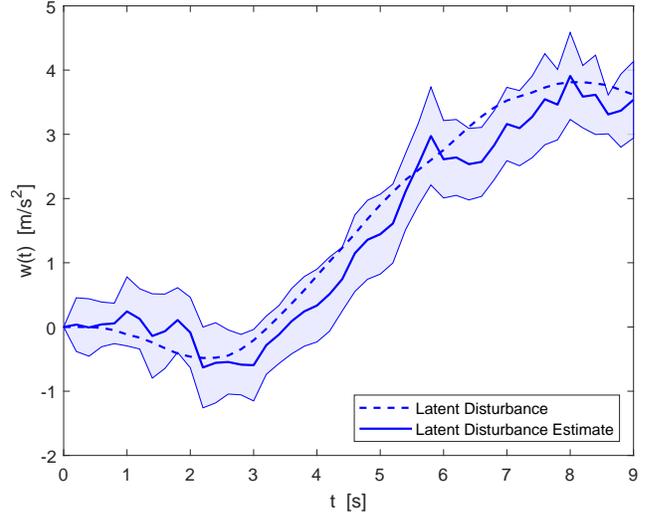}
\end{center}
\caption{Latent disturbance applied to the vehicle dynamics. The true latent disturbance is shown by the dashed blue line, and its estimate with $2\sigma$ bounds is shown by the solid blue line.}
\label{fig:Latent}
\end{figure}

\begin{figure}[h]
\begin{center}
\includegraphics[trim={4.25cm 8.5cm 4.25cm 8.75cm},clip,width=\linewidth]{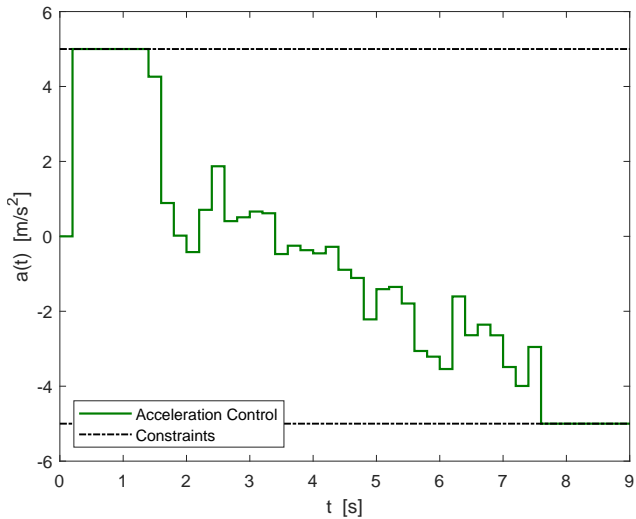}
\end{center}
\caption{The acceleration control applied to the vehicle.}
\label{fig:Accel}
\end{figure}

\begin{figure}[ht!]
\begin{center}
\includegraphics[trim={4.25cm 8.5cm 4.25cm 8.75cm},clip,width=\linewidth]{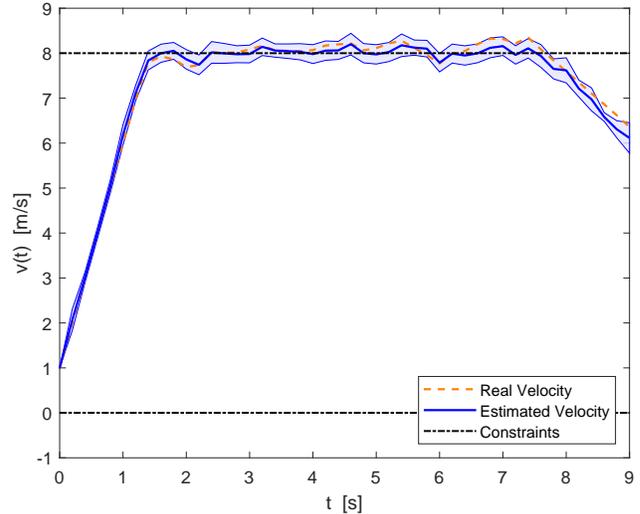}
\end{center}
\caption{Velocity profile of the vehicle. The true state is shown by the dashed orange line, and the estimated state with $2\sigma$ bounds is shown by the solid blue line.}
\label{fig:Velocity}
\end{figure}

The results of the motion planning example are shown in Figs. \ref{fig:Position}-\ref{fig:Velocity}. In Fig. \ref{fig:Position}, we see the resulting trajectory of the vehicle from the computed motion plan. The boundary constraints for the position of the car were met at every timestep during the simulation. The optimal trajectory performs as one would expect for the given constraints. The design leads to aggressive maneuvers during transitions between turns as observed, due in part to the chosen horizon length. The choice of the planning horizon must strike a balance between performance requirements, computation, and other practical factors.
We note the effect of the disturbance on the system and the controller response in Figs. \ref{fig:Latent} and \ref{fig:Accel}. The particle filter is able to accurately predict the underlying latent disturbance despite not having any direct measurements of it. The controller takes advantage of this information to counteract the effect of the latent force as depicted by the symmetry with the disturbance in Fig. \ref{fig:Accel}. The velocity profile, which is a direct result of the disturbance and acceleration control, is shown in Fig. \ref{fig:Velocity}. The velocity does struggle to satisfy the constraints at all time; however, from this figure we can see that the controller is able to keep the velocity near the bounds despite the very large disturbance with a maximum deviation of only $0.35$ m/s. We use a relatively low number of scenarios for satisfactory constraint satisfaction, while also optimizing for reduced conservatism. This lends to the efficacy of sampling-based optimization, even for nonlinear systems. A current downside is that the real-time applicability is limited by the computation demands of the controller, often taking much longer than the sampling time allows. 

\section{Conclusions} 
\label{Conclusions} 

Robotic systems often entail nonlinear uncertain dynamics and can be characterized as  nonlinear LFMs. A superior advantage of nonlinear LFMs lies in exploiting GPs to represent uncertainty and explicitly accounting for its presence in  the dynamic model.  In this paper, we investigated the problem of control design for temporal GP-based nonlinear LFMs. We used the equivalent state-space representation  of GP-based uncertainty  to convert a nonlinear LFM into an augmented state-space model and then formulated a stochastic MPC problem to control it.  The problem was then converted into a   deterministic problem via a scenario-based approach. The simulation results based on motion planning showed that the proposed approach is  effective   and that nonlinear LFMs are able to improve control performance by providing vital information about the disturbance. The disturbance could be more robustly modeled to capture state-dependent uncertainties by considering spatio-temporal GPs, which will be pursued in our future work. 


\balance
\bibliographystyle{IEEEtran}
\bibliography{NLFM_MPC}

\begin{thebibliography}{10}
\providecommand{\url}[1]{#1}
\csname url@samestyle\endcsname
\providecommand{\newblock}{\relax}
\providecommand{\bibinfo}[2]{#2}
\providecommand{\BIBentrySTDinterwordspacing}{\spaceskip=0pt\relax}
\providecommand{\BIBentryALTinterwordstretchfactor}{4}
\providecommand{\BIBentryALTinterwordspacing}{\spaceskip=\fontdimen2\font plus
\BIBentryALTinterwordstretchfactor\fontdimen3\font minus
  \fontdimen4\font\relax}
\providecommand{\BIBforeignlanguage}[2]{{%
\expandafter\ifx\csname l@#1\endcsname\relax
\typeout{** WARNING: IEEEtran.bst: No hyphenation pattern has been}%
\typeout{** loaded for the language `#1'. Using the pattern for}%
\typeout{** the default language instead.}%
\else
\language=\csname l@#1\endcsname
\fi
#2}}
\providecommand{\BIBdecl}{\relax}
\BIBdecl

\bibitem{Rosolia:ARCRAS:2018}
U.~Rosolia, X.~Zhang, and F.~Borrelli, ``Data-driven predictive control for
  autonomous systems,'' \emph{Annual Review of Control, Robotics, and
  Autonomous Systems}, vol.~1, no.~1, pp. 259--286, 2018.

\bibitem{Saltik:JPC:2018}
M.~B. Saltik, L.~Ozkan, J.~H. Ludlage, S.~Weiland, and P.~M. {Van den Hof},
  ``An outlook on robust model predictive control algorithms: Reflections on
  performance and computational aspects,'' \emph{Journal of Process Control},
  vol.~61, pp. 77 -- 102, 2018.

\bibitem{Bemporad:Springer:1999}
A.~Bemporad and M.~Morari, ``Robust model predictive control: A survey,'' in
  \emph{Robustness in Identification and Control}, A.~Garulli and A.~Tesi,
  Eds.\hskip 1em plus 0.5em minus 0.4em\relax London: Springer London, 1999,
  pp. 207--226.

\bibitem{Heirung:CCE:2018}
T.~A.~N. Heirung, J.~A. Paulson, J.~O’Leary, and A.~Mesbah, ``Stochastic
  model predictive control --- how does it work?'' \emph{Computers \& Chemical
  Engineering}, vol. 114, pp. 158 -- 170, 2018.

\bibitem{Zhang:AUTO:2020}
K.~Zhang and Y.~Shi, ``Adaptive model predictive control for a class of
  constrained linear systems with parametric uncertainties,''
  \emph{Automatica}, vol. 117, p. 108974, 2020.

\bibitem{Bujarbaruah:arXiv:2018}
M.~Bujarbaruah, X.~Zhang, H.~E. Tseng, and F.~Borrelli, ``Adaptive {MPC} for
  autonomous lane keeping,'' \emph{arXiv}, 2018, arXiv:1806.04335.

\bibitem{Rasmussen:MITPress:2005}
C.~E. Rasmussen and C.~K.~I. Williams, \emph{Gaussian Processes for Machine
  Learning (Adaptive Computation and Machine Learning)}.\hskip 1em plus 0.5em
  minus 0.4em\relax The MIT Press, 2005.

\bibitem{Klenske:TCST:2016}
E.~D. {Klenske}, M.~N. {Zeilinger}, B.~{Scholkopf}, and P.~{Hennig}, ``Gaussian
  process-based predictive control for periodic error correction,'' \emph{IEEE
  Transactions on Control Systems Technology}, vol.~24, no.~1, pp. 110--121,
  2016.

\bibitem{Ostafew:JFR:2016}
C.~J. Ostafew, A.~P. Schoellig, T.~D. Barfoot, and J.~Collier, ``Learning-based
  nonlinear model predictive control to improve vision-based mobile robot path
  tracking,'' \emph{Journal of Field Robotics}, vol.~33, no.~1, pp. 133--152,
  2016.

\bibitem{Carron:RAL:2019}
A.~{Carron}, E.~{Arcari}, M.~{Wermelinger}, L.~{Hewing}, M.~{Hutter}, and M.~N.
  {Zeilinger}, ``Data-driven model predictive control for trajectory tracking
  with a robotic arm,'' \emph{IEEE Robotics and Automation Letters}, vol.~4,
  no.~4, pp. 3758--3765, 2019.

\bibitem{Hewing:TCST:2020}
L.~{Hewing}, J.~{Kabzan}, and M.~N. {Zeilinger}, ``Cautious model predictive
  control using {Gaussian} process regression,'' \emph{IEEE Transactions on
  Control Systems Technology}, vol.~28, no.~6, pp. 2736--2743, 2020.

\bibitem{Kocijan:ACC:2004}
J.~{Kocijan}, R.~{Murray-Smith}, C.~E. {Rasmussen}, and A.~{Girard}, ``Gaussian
  process model based predictive control,'' in \emph{Proceedings of the 2004
  American Control Conference}, vol.~3, 2004, pp. 2214--2219.

\bibitem{Cao:JIRS:2017}
G.~Cao, E.~M.-K. Lai, and F.~Alam, ``Gaussian process model predictive control
  of an unmanned quadrotor,'' \emph{Journal of Intelligent {\&} Robotic
  Systems}, vol.~88, no.~1, pp. 147--162, Oct 2017.

\bibitem{Yang:SYSID:2015}
X.~Yang and J.~Maciejowski, ``Risk-sensitive model predictive control with
  gaussian process models,'' in \emph{Proceedings of 17th IFAC Symposium on
  System Identification}, vol.~48, no.~28, 2015, pp. 374 -- 379, sYSID 2015.

\bibitem{Kim:IFAC:2017}
T.~Kim, W.~Kim, S.~Choi, and H.~{Jin Kim}, ``Path tracking for a skid-steer
  vehicle using model predictive control with on-line sparse gaussian
  process,'' in \emph{Proceedings of the 20th IFAC World Congress}, vol.~50,
  no.~1, 2017, pp. 5755 -- 5760.

\bibitem{Sarkka:TAC:2019}
S.~S{\"a}rkk{\"a}, M.~A. {\'{A}lvarez}, and N.~D. {Lawrence}, ``Gaussian
  process latent force models for learning and stochastic control of physical
  systems,'' \emph{IEEE Transactions on Automatic Control}, vol.~64, no.~7, pp.
  2953--2960, 2019.

\bibitem{Grasshoff:CDC:2019}
J.~Gra{\ss}hoff, G.~M{\"a}nnel, H.~S. Abbas, and P.~Rostalksi, ``Model
  predictive control using efficient gaussian processes for unknown disturbance
  inputs,'' in \emph{Proceedings of the IEEE 58th Conference on Decision and
  Control (CDC)}, 2019, pp. 2708--2713.

\bibitem{Hartikainen:MLSP:2010}
J.~{Hartikainen} and S.~{Särkkä}, ``Kalman filtering and smoothing solutions
  to temporal gaussian process regression models,'' in \emph{2010 IEEE
  International Workshop on Machine Learning for Signal Processing}, 2010, pp.
  379--384.

\bibitem{Sarkka:MLSP:2014}
S.~{Särkkä} and R.~{Piché}, ``On convergence and accuracy of state-space
  approximations of squared exponential covariance functions,'' in \emph{2014
  IEEE International Workshop on Machine Learning for Signal Processing}, 2014,
  pp. 1--6.

\bibitem{Hartikainen:ICML:2012}
J.~{Hartikainen}, M.~{Sepp\"{a}nen}, and S.~{S\"{a}rkk\"{a}}, ``State-space
  inference for non-linear latent force models with application to satellite
  orbit prediction,'' in \emph{Proceedings of the 29th International Coference
  on International Conference on Machine Learning}, ser. ICML'12.\hskip 1em
  plus 0.5em minus 0.4em\relax Madison, WI, USA: Omnipress, 2012, p. 723–730.

\bibitem{Zhao:arxiv:2020}
Z.~Zhao, T.~Karvonen, R.~Hostettler, and S.~Särkkä, ``Taylor moment expansion
  for continuous-discrete gaussian filtering and smoothing,'' \emph{arXiv},
  2020, arXiv:2001.02466.

\bibitem{Andrieu:bookch:2001}
C.~Andrieu, A.~Doucet, and E.~Punskaya, \emph{Sequential Monte Carlo Methods
  for Optimal Filtering}.\hskip 1em plus 0.5em minus 0.4em\relax New York, NY:
  Springer New York, 2001, pp. 79--95.

\bibitem{Calafiore:TAC:2006}
G.~C. {Calafiore} and M.~C. {Campi}, ``The scenario approach to robust control
  design,'' \emph{IEEE Transactions on Automatic Control}, vol.~51, no.~5, pp.
  742--753, 2006.

\bibitem{Bernardini:CDC:2009}
D.~{Bernardini} and A.~{Bemporad}, ``Scenario-based model predictive control of
  stochastic constrained linear systems,'' in \emph{Proceedings of the 48h IEEE
  Conference on Decision and Control}, 2009, pp. 6333--6338.

\bibitem{Schildbach:AUTO:2014}
G.~Schildbach, L.~Fagiano, C.~Frei, and M.~Morari, ``The scenario approach for
  stochastic model predictive control with bounds on closed-loop constraint
  violations,'' \emph{Automatica}, vol.~50, no.~12, pp. 3009 -- 3018, 2014.

\bibitem{Shang:JPC:2019}
C.~Shang and F.~You, ``A data-driven robust optimization approach to
  scenario-based stochastic model predictive control,'' \emph{Journal of
  Process Control}, vol.~75, pp. 24 -- 39, 2019.

\bibitem{Kang:ACC:2015}
J.~{Kang}, A.~U. {Raghunathan}, and S.~{Di Cairano}, ``Decomposition via {ADMM}
  for scenario-based model predictive control,'' in \emph{Proceedings of the
  American Control Conference}, 2015, pp. 1246--1251.

\bibitem{Sehr:IFAC:2017}
M.~A. Sehr and R.~R. Bitmead, ``Particle model predictive control: Tractable
  stochastic nonlinear output-feedback {MPC},'' \emph{Proceedings of 20th IFAC
  World Congress}, vol.~50, no.~1, pp. 15\,361 -- 15\,366, 2017, 20th IFAC
  World Congress.

\bibitem{Campi:SIAM:2008}
M.~C. Campi and S.~Garatti, ``The exact feasibility of randomized solutions of
  uncertain convex programs,'' \emph{SIAM Journal on Optimization}, vol.~19,
  no.~3, pp. 1211--1230, 2008.

\bibitem{Rajamani:book:2011}
R.~Rajamani, \emph{Vehicle Dynamics and Control}, ser. Mechanical Engineering
  Series.\hskip 1em plus 0.5em minus 0.4em\relax Springer US, 2011.

\bibitem{MATLAB:2019}
MatLab, ``fmincon,'' in \emph{Optimization Toolbox}.\hskip 1em plus 0.5em minus
  0.4em\relax Natick, Massachusetts: The MathWorks Inc., 2019.

\end{thebibliography}


\end{document}